\documentclass[acmtog, screen, nonacm]{acmart}

\usepackage{booktabs} 
\usepackage{bbm}
\usepackage{nicefrac}
\usepackage{algorithm}
\usepackage{algpseudocode}
\usepackage[normalem]{ulem}
\usepackage{empheq} 
\usepackage{algorithm}
\usepackage{algpseudocode}
\usepackage{hyperref}
\setcopyright{none}
\setcitestyle{square}

\title{Dress-1-to-3: Single Image to Simulation-Ready 3D Outfit with \\ Diffusion Prior and Differentiable Physics}

\author{Xuan Li}
\authornotemark[1]
\affiliation{
\institution{UCLA}
}

\author{Chang Yu}
\authornotemark[1]
\affiliation{
\institution{UCLA}
}

\author{Wenxin Du}
\authornotemark[1]
\affiliation{
\institution{UCLA}
}

\author{Ying Jiang}
\authornote{\ indicates equal contributions.}
\affiliation{
\institution{UCLA}
}

\author{Tianyi Xie}
\affiliation{
\institution{UCLA}
}

\author{Yunuo Chen}
\affiliation{
\institution{UCLA}
}

\author{Yin Yang}
\affiliation{
\institution{University of Utah}
}

\author{Chenfanfu Jiang}
\affiliation{
\institution{UCLA}
}

\makeatletter
\let\@authorsaddresses\@empty
\makeatother

\makeatletter
\def\@ACM@checkaffil{%
    \if@ACM@instpresent\else
    \ClassWarningNoLine{\@classname}{No institution present for an affiliation}%
    \fi
    \if@ACM@citypresent\else
    \ClassWarningNoLine{\@classname}{No city present for an affiliation}%
    \fi
    \if@ACM@countrypresent\else
    \ClassWarningNoLine{\@classname}{No country present for an affiliation}%
    \fi
}
\makeatother

\usepackage{algorithm}
\usepackage{amsmath}

\DeclareMathOperator*{\argmin}{arg\,min}
\usepackage{algorithmicx}
\usepackage{algpseudocode}
\usepackage{multirow}
\usepackage{caption}
\usepackage{subcaption,bm}
\usepackage{ stmaryrd }
\usepackage{wrapfig}
\algdef{SE}[DOWHILE]{Do}{doWhile}{\algorithmicdo}[1]{\algorithmicwhile\ #1}
\AtBeginDocument{%
  \providecommand\BibTeX{{%
    \normalfont B\kern-0.5em{\scshape i\kern-0.25em b}\kern-0.8em\TeX}}}

\newcommand{\dd}{\text{d}}

\begin{document}

\begin{abstract}
Recent advances in large models have significantly advanced image-to-3D reconstruction. However, the generated models are often fused into a single piece, limiting their applicability in downstream tasks. This paper focuses on 3D garment generation, a key area for applications like virtual try-on with dynamic garment animations, which require garments to be separable and simulation-ready. We introduce Dress-1-to-3, a novel pipeline that reconstructs physics-plausible, simulation-ready separated garments with sewing patterns and humans from an in-the-wild image. Starting with the image, our approach combines a pre-trained image-to-sewing pattern generation model for creating coarse sewing patterns with a pre-trained multi-view diffusion model to produce multi-view images. The sewing pattern is further refined using a differentiable garment simulator based on the generated multi-view images. Versatile experiments demonstrate that our optimization approach substantially enhances the geometric alignment of the reconstructed 3D garments and humans with the input image. Furthermore, by integrating a texture generation module and a human motion generation module, we produce customized physics-plausible and realistic dynamic garment demonstrations. Our project page is \url{https://dress-1-to-3.github.io/}.
\end{abstract}

\begin{teaserfigure}
  \includegraphics[width=\textwidth]{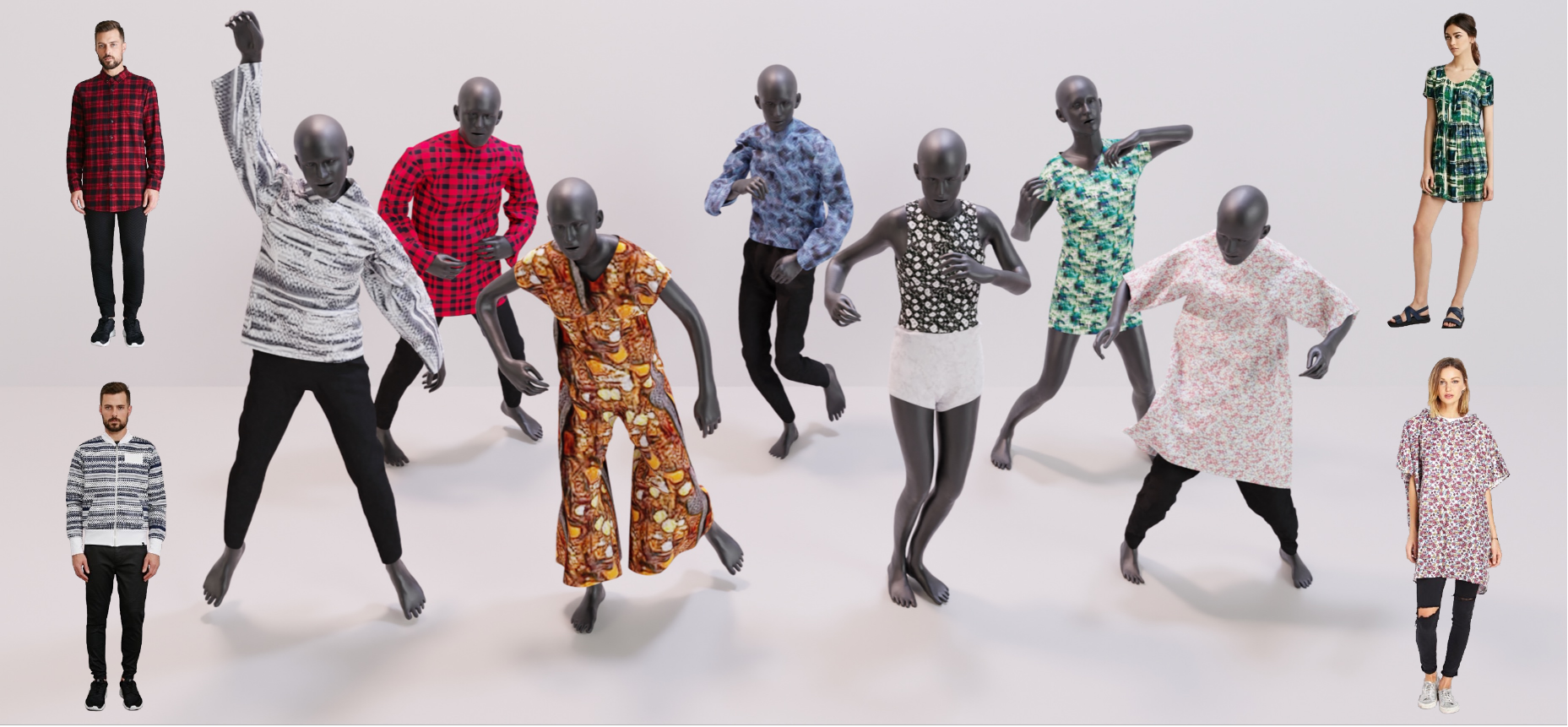}
  \caption{Dress-1-to-3 can reconstruct simulation-ready textured clothed humans from casually posed single view images.}
  \label{fig:teaser}
\end{teaserfigure}

\maketitle
\pagestyle{plain}

\section{Introduction}
Creating digital assets of clothed humans is crucial for a wide range of applications, including virtual reality (VR), the film industry, fashion design, and gaming. However, the traditional pipeline for digital human and garment creation involves multiple intricate steps, such as concept design, material selection, garment modeling, human pose generation, garment fitting, and animation. These processes are often labor-intensive and time-consuming. 

In recent years, significant advancements in image-to-3D asset reconstruction have been driven by the development of powerful image and video generation models. Among these, multiview diffusion models \cite{chen2024v3d, liu2023zero, gao2024cat3d} have emerged as a promising approach, effectively leveraging multiview images as intermediate representations to capture 3D information. When fine-tuned on human datasets, these models generalize well to avatar reconstructions from in-the-wild images \cite{li2024pshuman, he2024magicman}. However, the generated results are often fused into a single piece, making them unsuitable for downstream tasks such as garment animation and interaction. 

In the meantime, sewing patterns, a foundational representation in the garment design industry, have been adopted as intermediate reconstruction outputs to recover garment geometries \cite{liu2023towards, li2024garment}. This representation is particularly advantageous due to its seamless integration with downstream applications such as physics simulation and garment editing. Despite their promise, these feed-forward approaches face significant limitations stemming from the scarcity of high-quality 3D data. As a result, the reconstructed garments are often constrained by the distribution of the training dataset, leading to inaccuracies in aligning with input images. This limitation hinders their ability to produce detailed and diverse reconstructions reflective of real-world garment variations. The question then arises: can we keep the advantages of the simulation-ready representation of sewing patterns while leveraging the powerful priors in large multi-view diffusion models to reconstruct garments from solely an in-the-wild image?

To address this problem, we introduce \textbf{Dress-1-to-3}, a novel garment reconstruction pipeline that accurately transforms an in-the-wild image into a simulation-ready representation of separated human and garment by leveraging the strengths of both 2D multi-view diffusion and 3D sewing pattern reconstruction. To bridge those two parts, we propose a generalized and unified IPC differentiable framework for garment optimization, which enables the optimization of 3D sewing patterns using 2D generative multi-view RGB images and normal maps as guidance. By refining imperfect generative outputs to align with the geometry encoded in multiview images, our approach allows the reconstruction of out-of-distribution garment shapes with high fidelity. Our contributions include: 

\begin{itemize}
    \item We propose a holistic garment reconstruction pipeline that takes a single image as input and generates garments fitted onto a posed human, ensuring both the human pose and garments align with the input image.
    \item We derive a generalized and unified IPC differentiable framework that is agonistic to constitutive models. We apply this framework for co-dimensional garment optimization.
    \item We conduct extensive experiments to demonstrate the effectiveness and versatility of our garment optimization framework, successfully reconstructing garments across diverse categories, including those not present in the training dataset.
\end{itemize}

\section{Related Work}

\subsection{Multi-view Diffusion} 

Owing to their powerful predictive ability, Diffusion Probabilistic Models \cite{ho2020denoising} have been applied to image \cite{nichol2021glide, zhang2023adding, dhariwal2021diffusion, ruiz2023dreambooth, saharia2022palette}, video \cite{chen2024videocrafter2, ho2022video}, and 3D shape synthesis tasks \cite{long2024wonder3d, yu2024surf, tang2023dreamgaussian}, etc. However, applying image diffusion models to generate multi-view images separately poses significant challenges in maintaining consistency across different views. To address multi-view inconsistency, multi-view attentions and camera pose controls are adopted to fine-tune pre-trained image diffusion models, enabling the simultaneous synthesis of multi-view images \cite{shi2023mvdream, wang2023imagedream, xu2023dmv3d, yang2024consistnet, shi2023zero123++, long2024wonder3d}, though these methods might result in compromised geometric consistency due to the lack of inherent 3D biases. To ensure both global semantic consistency and detailed local alignment in multi-view diffusion models, 3D-adapters \cite{chen20243d} propose a plug-in module designed to infuse 3D geometry awareness. Nevertheless, the generated images by these models are sparse views. To address this issue, CAT3D \cite{gao2024cat3d} introduces an efficient parallel sampling strategy to generate a large set of camera poses, and MVDiffusion++ \cite{tang2025mvdiffusion++} adopts a pose-free architecture and a view dropout strategy to reduce computational costs, generating dense, high-resolution images.

Generating consistent images from multi-view diffusions offers guidance for further 3D shape reconstruction \cite{gao2024cat3d}. PSHuman \cite{li2024pshuman} integrates a body-face cross-scale diffusion with an SMPL-X conditioned multi-view diffusion for clothed human reconstruction with high-quality face details. Recent work, MagicMan \cite{he2024magicman}, utilizes a hybrid human-specific multi-view diffusion model with 3D SMPL-X-based body priors and 2D diffusion priors to consistently generate dense multi-view RGB images and normal maps, supporting high-quality human mesh reconstruction. Different from these works, we exploited multi-view diffusions to generate multi-view normals and RGB images as guidance to optimize sewing patterns and stitches instead of human meshes.

\subsection{Garment Reconstruction} 

Previous work focusing on clothed human reconstruction \cite{xiu2022icon, xiu2023econ} typically generates garments fused with digital human models, limiting them to basic skinning-based animations and requiring extra segmentation and editing to separate the garments from the human body. In contrast, our approach focuses on reconstructing separately wearable, simulation-ready garments and human models. Other closely related works include \citet{li2024diffavatar, yu2024inverse}, which also generates simulation-ready clothes via differentiable simulation, but at the cost of creating clothing templates by artists, precise point clouds by scanners or 3D shapes of garments. NeuralTailor \cite{korosteleva2022neuraltailor} utilizes point-level attention for pattern shape and stitching information regression, enabling the reconstruction of garment meshes from point clouds. In contrast, our paper focuses on reconstructing non-watertight garments and humans separately from a single image without additional inputs. 

To reconstruct separated non-watertight garments from a single image, GarVerseLOD \cite{luo2024garverselod} recovers garment details hierarchically in a coarse-to-fine framework. However, it fails to reconstruct complex skirts or dresses with slits or with complex human poses due to the limited representation of such features in the training data. ClothWild \cite{moon20223d} exploits a weakly supervised pipeline with DensePose-based loss to further increase robustness on in-the-wild images. BCNet \cite{jiang2020bcnet} introduces a layered garment representation and a generic skinning weight generation network to model garments with different topologies. Deep Fashion3D \cite{zhu2020deep} refines adaptable templates with rich annotations to fit garment shapes. While they are limited to garment categories in their training datasets, these works fail to reconstruct complex categories such as jumpsuits. Additionally, they require nearly frontal images as input, limiting reconstruction from different views. AnchorUDF \cite{zhao2021learning} explores a learnable unsigned distance function to query both 3D position features and pixel-aligned image features via anchor points, which reconstructs the coarse garment shape but lacks the generation of high-quality geometric details. 

Instead of directly reconstructing garment meshes, some works \cite{liu2023towards, he2024dresscode, korosteleva2023garmentcode} treat sewing patterns as intermediate representations to generate garments by stitching them together. Recent work, GarmentRecovery \cite{li2024garment}, introduces implicit sewing patterns (ISP) to provide shape priors integrated with deformation priors for further garment recovery, though it builds specialized models for each individual garment or garment type. Both SewFormer \cite{liu2023towards}, and PanelFormer \cite{chen2024panelformer} utilize Transformers to predict sewing patterns and stitches. Concurrent work, SewingLDM, AIpparel, Design2GarmentCode, and ChatGarment \cite{liu2024multimodal, nakayama2024aipparel, zhou2024design2garmentcode, bian2024chatgarment}, exploits multimodal models to synthesize sewing patterns. However, their garment results lack physical material parameters. Therefore, they fail to reconstruct diverse shapes for garments with different physical materials. While \citet{wang2018learning} and \citet{yang2018physics} leverage a shared latent space and joint material-pose optimization to generate 3D garments and 2D sewing patterns, their approaches rely heavily on large-scale datasets on garment templates and human-body models, limiting their ability to generalize to out-of-distribution garments and body shapes. Our work aims to generate diverse, image-aligned, simulation-ready garments with high-quality details from in-the-wild images by optimizing sewing patterns and stitches with physical parameters via differentiable simulations. 

\subsection{Differentiable Simulation}

Differentiable simulation has seen widespread application in recent research, particularly for system identification and the inference of material parameters from both synthetic~\cite{li2023pac, li2024neuralfluid} and real-world~\cite{huang2024differentiable, si2024difftactile} observations. The scope of exploration spans various domains, including fluid dynamics and control~\cite{mcnamara2004fluid, schenck2018spnets, li2023difffr, li2024neuralfluid}, rigid-body dynamics~\cite{freeman2021brax, strecke2021diffsdfsim, xu2023efficient}, articulated systems~\cite{geilinger2020add, qiao2021efficient, xu2021end}, soft-body dynamics~\cite{hahn2019real2sim, hu2019chainqueen, du2021diffpd, jatavallabhula2021gradsim, huang2024differentiable}, cloth~\cite{li2022diffcloth, stuyck2023diffxpbd, li2024diffavatar}, inelasticity~\cite{huang2021plasticinelab, li2023pac}, inflatable structures~\cite{panetta2021computational}, and Voronoi diagrams~\cite{numerow2024differentiable}.

Cloth-based applications, whether for static optimization or dynamic simulation \cite{santesteban2022snug, grigorev2023hood}, frequently involve extensive frictional contact. Consequently, many works focus on robust methods for handling dry frictional contact in differentiable simulations. \citet{bartle2016physics} proposes a physics-driven pattern adjustment for garment editing using fixed-point optimization, which does not account for gradients. \citet{liang2019differentiable} is the first to introduce a fully functional differentiable cloth simulator with frictional contact and self-collision, formulating a quadratic programming problem. \citet{jatavallabhula2021gradsim} employs a penalty-based frictional contact model, while \citet{du2021diffpd} and \citet{li2022diffcloth} leverage the adjoint method for Projective Dynamics~\cite{bouaziz2014projective} with friction. Building on Position-Based Dynamics~\cite{muller2007position, macklin2016xpbd}, \citet{stuyck2023diffxpbd} and \citet{li2024diffavatar} introduce differentiable formulations for compliant constraint dynamics, and \citet{huang2024differentiable} presents an adjoint-based framework for differentiable Incremental Potential Contact (IPC)~\cite{li2020incremental, li2020codimensional}.

The finite difference (FD) method~\cite{renardy2006introduction} is a standard approach to numerical differentiation. The complex-step finite difference technique~\cite{luo2019accelerated, shen2021high} offers an alternative that mitigates issues such as subtractive cancellation and accumulated numerical errors by leveraging complex Taylor expansions~\cite{brezillon1981numerical}. They can be used to optimize low-DoF system \cite{zheng2025physavatar}. Automatic differentiation (AD)~\cite{naumann2011art, margossian2019review} and code transformation libraries like NVIDIA Warp~\cite{macklin2022warp}, DiffTaichi~\cite{hu2019chainqueen, hu2019difftaichi}, and others~\cite{herholz2024mesh} automatically compute gradients based on forward simulation, allowing for greater reuse of existing code. However, they can introduce code constraints, incur a high memory footprint, and may cause gradient explosion if applied naively. Our framework combines NVIDIA Warp’s AD with an adjoint method to achieve both development efficiency and high performance.

\section{Differentiable Garment Simulation} \label{sec:sim}
\subsection{Forward Simulation}
We use Codimensional Incremental Potential Contact (CIPC) \citep{li2020codimensional} as our underlying garment simulation method, which is the state-of-the-art in cloth simulation regarding accuracy and robustness. It ensures non-penetration through distance-based log barrier energy and continuous collision detection (CCD). Below, we summarize the simulation pipeline, with further details available in \citet{li2020codimensional}.

The simulated codimensional surface is discretized into triangles defined by vertices $\bm V$ and faces $\bm F$. Let $\bm X$ denote the vertex positions in the undeformed state, and let $\bm x^n$ and $\bm v^n$ represent the vertex positions and velocities, respectively, at time step $t^n$. CIPC employs an optimization-based time integrator to achieve the state transition from time step $t^n$ to $t^{n+1} = t^n + h$, minimizing the following energy:
\begin{equation}
\bm x^{n+1} = \argmin_{\bm x} E(\bm x) = \frac{1}{2} \|\bm x - \tilde{\bm x}\|_{\bm M}^2 + \Psi(\bm x; \bm X) + B(\bm x).
\label{eq:ipc_optimization}
\end{equation}
Here, $\tilde{\bm x} = \bm x^n + \bm v^n h + \bm g h^2$ represents the predictive position under backward Euler integration. $\|\cdot\|_{\bm M}$ denotes the $L^2$-norm weighted by the vertex mass $\bm M_{ii}$. $\Psi(\bm x; \bm X)$ is the elastic energy, encompassing both stretching and bending energies, depending on the user's choice. $B(\bm x)$ is the log barrier energy introduced by IPC, defined over all contacting vertex-triangle and edge-edge pairs. The barrier energy for each pair of primitives increases from zero to infinity as the gap decreases from a threshold $\hat{d}$ to $0$, providing sufficient repulsion to prevent penetrations.

Newton's method with line search is employed to solve the optimization problem, requiring the analytical computation of the gradient and Hessian matrix of the energy at each iteration. The step size upper bound in each line search is clamped by CCD to ensure that all intermediate states remain intersection-free, provided that $\bm x^n$ is initially intersection-free. Finally, the new velocity is updated as $\bm v^{n+1} = (\bm x^{n+1} - \bm x^n) / h$.

\subsection{Differentiable CIPC}

\citet{huang2024differentiable} provided an analytical derivation of differentiable IPC using the adjoint method. However, their derivation is closely tied to specific choices of constitutive models. To extend their framework to support cloth simulation, tedious derivations of analytical derivatives are required. In this work, we present a simple and unified framework that leverages both automatic differentiation and the adjoint method.

The governing equation of CIPC simulation can be expressed as an implicit nonlinear system of equations derived from the first-order optimality condition of the minimizer for \autoref{eq:ipc_optimization}:
\begin{align}
    &\bm G(\bm x^*; \bm x^n, \bm v^n, \bm \varsigma^n) = \nabla E(\bm x^*; \bm x^n, \bm v^n, \bm \varsigma^n) = \bm 0,  \label{eq:implicit}\\
    &\bm x^{n+1} = \bm x^*,  \quad \bm v^{n+1} = \frac{1}{h}(\bm x^* - \bm x^{n}), \label{eq:update}
\end{align}
Here, $\bm x^*$ is the minimizer of the system energy $E$, ${\bm x^n, \bm v^n}$ represents the last system state, and $\bm \varsigma^n$ denotes the set of all continuous parameters of the implicit equation, including shape parameters $\bm X$, mass matrix $\bm M$, elastic moduli, and others. We assume ${\bm \varsigma^n}$ are independent, although they may share the same values. This abstraction allows the simulator to function as a differentiable layer with ${\bm x^{n}, \bm v^{n}, \bm \varsigma^n}$ as input and ${\bm x^{n + 1}, \bm v^{n+1}}$ as output. The computational graph can be handled by any auto-differentiable framework such as PyTorch. The backward operator computes $\frac{\dd\mathcal{L}}{\dd \bm x^{n}}$, $\frac{\dd\mathcal{L}}{\dd \bm v^{n}}$, and $\frac{\dd\mathcal{L}}{\dd \bm \varsigma^{n}}$ given $\frac{\dd\mathcal{L}}{\dd \bm x^{n+1}}$ and $\frac{\dd\mathcal{L}}{\dd \bm v^{n+1}}$ for a given training loss function $\mathcal{L}$.

Taking the full derivatives of \autoref{eq:implicit} with respect to ${\bm x^n, \bm v^n, \bm \varsigma^n}$ on both sides, we obtain:
\begin{equation}
  \frac{\partial \bm G}{\partial \bm x^{*}}\left[\frac{\dd \bm x^{*}}{\dd \bm x^n}, \frac{\dd \bm x^{*}}{\dd \bm v^n}, \frac{\dd \bm x^{*}}{\dd \bm \varsigma^n}\right] + \left[\frac{\partial \bm G}{\partial \bm x^n}, \frac{\partial G}{\partial \bm v^n}, \frac{\partial G}{\partial \bm \varsigma^n}\right] = \bm 0,
\end{equation}
which leads to 
\begin{equation}
    \left[\frac{\dd \bm x^{*}}{\dd \bm x^n}, \frac{\dd \bm x^{*}}{\dd \bm v^n}, \frac{\dd \bm x^{*}}{\dd \bm \varsigma^n}\right] = -\left[\frac{\partial \bm G}{\partial \bm x^{*}}\right]^{-1} \left[\frac{\partial \bm G}{\partial \bm x^n}, \frac{\partial G}{\partial \bm v^n}, \frac{\partial G}{\partial \bm \varsigma^n}\right] \label{eq:adjoint}.
\end{equation}
By the chain rule, we have:
\begin{equation}
\begin{split}
    \left[\frac{\dd \mathcal{L}}{\dd \bm x^n}, \frac{\dd \mathcal{L}}{\dd \bm v^n}, \frac{\dd \mathcal{L}}{\dd \bm \varsigma^n}\right] =  \frac{\dd\mathcal{L}}{\dd\bm x^{n+1}} \left[\frac{\dd \bm x^{n+1}}{\dd \bm x^n}, \frac{\dd \bm x^{n+1}}{\dd \bm v^n}, \frac{\dd \bm x^{n+1}}{\dd \bm \varsigma^n}\right]& \\
      +  \frac{\dd\mathcal{L}}{\dd\bm v^{n+1}} \left[\frac{\dd \bm v^{n+1}}{\dd \bm x^n}, \frac{\dd \bm v^{n+1}}{\dd \bm v^n}, \frac{\dd \bm v^{n+1}}{\dd \bm \varsigma^n}\right].&
\end{split}
\label{eq:chain_rule}
\end{equation}
Here, we assume $\frac{\dd \mathcal{L}}{\dd \bm x^n}$, $\frac{\dd \mathcal{L}}{\dd \bm v^n}$, and $\frac{\dd \mathcal{L}}{\dd \bm \varsigma^n}$ are all row vectors to ensure dimension consistency. From \autoref{eq:update}, we have:
\begin{equation}
\dd \bm  x^{n+1} = \dd \bm x^{*},\quad \dd \bm  v^{n+1} = \frac{1}{h}(\dd \bm x^{*} - \dd \bm x^n). 
\label{eq:eq3_diff}
\end{equation}
Plugging \autoref{eq:eq3_diff} into \autoref{eq:chain_rule}, we obtain:
\begin{equation}
\begin{split}
   \left[\frac{\dd \mathcal{L}}{\dd \bm x^n}, \frac{\dd \mathcal{L}}{\dd \bm v^n}, \frac{\dd \mathcal{L}}{\dd \bm \varsigma^n}\right] = \frac{\dd\mathcal{L}}{\dd\bm x^{n+1}} \left[\frac{\dd \bm x^{*}}{\dd \bm x^n}, \frac{\dd \bm x^{*}}{\dd \bm v^n}, \frac{\dd \bm x^{*}}{\dd \bm \varsigma^n}\right] &\\
     + \frac{1}{h} \frac{\dd\mathcal{L}}{\dd\bm v^{n+1}} \left[\frac{\dd \bm x^*}{\dd \bm x^n} - \bm I, \frac{\dd  \bm x^*}{\dd \bm v^n}, \frac{\dd  \bm x^*}{\dd \bm \varsigma^n}\right].&
\end{split}
\end{equation}
With some rearrangements, we arrive at:
\begin{equation}
\begin{split}
    &\frac{\dd\mathcal{L}}{\dd  \bm x^n} = \left[\frac{\dd\mathcal{L}}{\dd\bm{x}^{n+1}} + \frac{1}{h}\frac{\dd\mathcal{L}}{\dd \bm{v}^{n+1}} \right]\frac{\dd\bm x^{*}}{\dd\bm x^n} - \frac{1}{h}\frac{\dd\mathcal{L}}{\dd \bm{v}^{n+1}} \\
    &\left[\frac{\dd\mathcal{L}}{\dd \bm v^n}, \frac{\dd\mathcal{L}}{\dd \bm \varsigma^n}\right] = \left[\frac{\dd\mathcal{L}}{\dd\bm{x}^{n+1}} + \frac{1}{h}\frac{\dd\mathcal{L}}{\dd \bm{v}^{n+1}} \right] \left[\frac{\dd \bm x^{*}}{\dd \bm v^n}, \frac{\dd \bm x^{*}}{\dd \bm \varsigma^n}\right].
\end{split}
\end{equation}
Denote $\mathcal{A} = \left[\frac{\dd\mathcal{L}}{\dd\bm{x}^{n+1}} + \frac{1}{h}\frac{\dd\mathcal{L}}{\dd \bm{v}^{n+1}}\right]\left[\frac{\partial \bm G}{\partial \bm x^{*}}\right]^{-1}$. By \autoref{eq:adjoint}, we have:
\begin{equation}
 \frac{\dd\mathcal{L}}{\dd  \bm x^n} = - \mathcal{A} \frac{\partial \bm G}{\partial \bm x^n}  - \frac{1}{h}\frac{\dd\mathcal{L}}{\dd \bm{v}^{n+1}},
\end{equation}
\begin{equation}
\left[\frac{\dd\mathcal{L}}{\dd \bm v^n}, \frac{\dd\mathcal{L}}{\dd \bm \varsigma^n}\right] = - \mathcal{A} \left[\frac{\partial \bm G}{\partial \bm v^n}, \frac{\partial \bm G}{\partial \bm \varsigma^n}\right].
\end{equation}
Observe that $\mathcal{A}$ is obtained by solving a linear system, where the coefficient matrix $\frac{\partial \bm G}{\partial \bm x^{*}}$ is the Hessian matrix of the system energy $E$. The term $\mathcal{A} \left[\frac{\partial \bm G}{\partial \bm x^n}, \frac{\partial \bm G}{\partial \bm v^n}, \frac{\partial \bm G}{\partial \bm \varsigma^n}\right]$ back-propagates the differentials in $\mathcal{A}$ to $\bm x^n$, $\bm v^n$, and $\bm \varsigma^n$ through $\bm G$, respectively. This process can be implemented by treating $\bm G$ as a differentiable layer that supports auto-differentiation. Using AutoDiff, we eliminate the need to manually derive the analytical expressions for $\frac{\partial \bm G}{\partial \bm v^n}$ and $\frac{\partial \bm G}{\partial \bm \varsigma^n}$. All other components required for forward simulations have already been derived.

\begin{figure*}
\includegraphics[width=\textwidth]{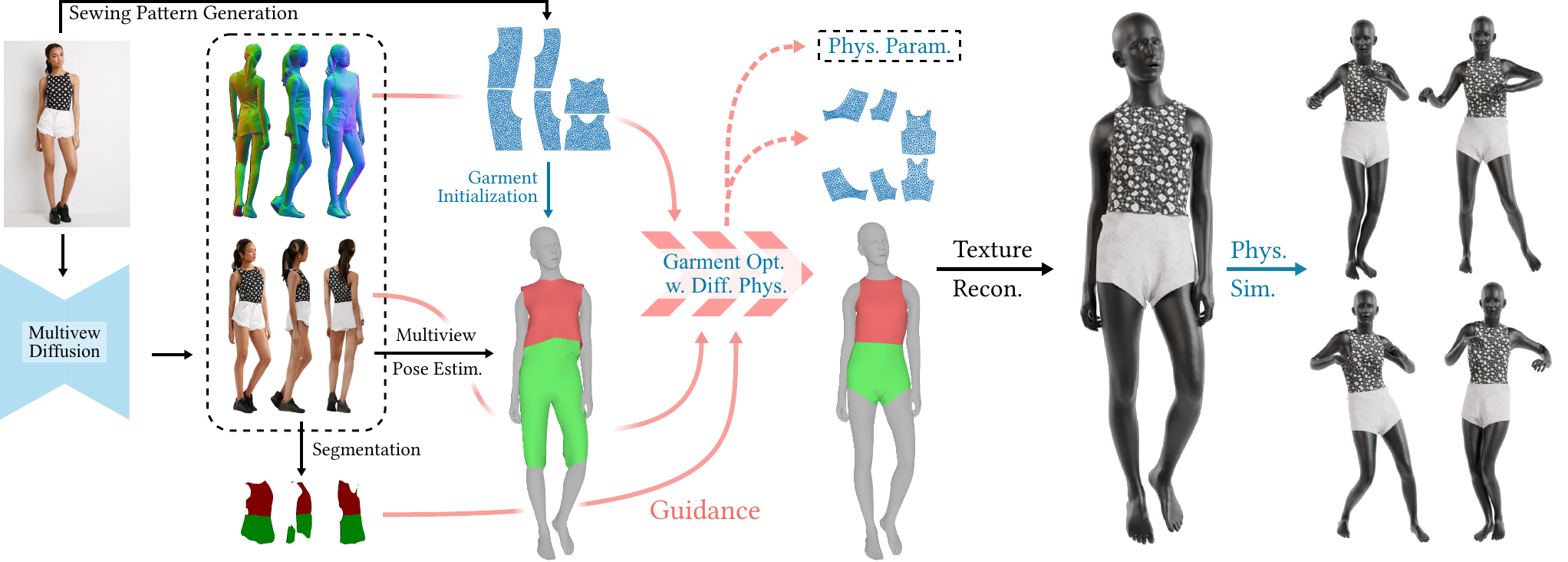}
\caption{\textbf{Dress-1-to-3 Pipeline.} Starting with a single-view input image of a clothed human, we first derive an initial estimation of the sewing pattern. Additionally, we employ multi-view diffusion to generate orbital camera views, which serve as ground-truth 3D information for both human pose and garment shape. Next, we utilize differentiable simulation to sew and drape the pattern onto the posed human model, optimizing its shape and physical parameters in conjunction with geometric regularizers. Finally, the optimized garment shape provides a physically plausible rest shape in its static state and is readily animatable using a physical simulator.}
\label{fig:pipeline}
\end{figure*}

\section{Method Overview}
We start our pipeline by estimating an initial garment sewing pattern from a single-view image. Next, we generate consistent multi-view RGB images and their corresponding normal maps, based on which we predict the human body pose. The 3D garment is initialized by stitching and draping the 2D patterns onto the predicted human model. The garment's interaction with the human body is simulated using a differentiable CIPC simulator, allowing us to optimize the physical parameters and the shapes of the sewing patterns guided by the previously generated multi-view RGB images, normal maps, and segmentation results. The optimized state produces a simulation-ready scene with a human model wearing well-fitted 3D outfits that align with the input. Garment textures are automatically generated using a visual-language model and image diffusion. Finally, by applying our CIPC simulator, we can simulate dynamic scenes where the predicted human body wears the optimized garments while performing various motion sequences. An illustration of the pipeline is shown in \autoref{fig:pipeline}. We elaborate on each component of the pipeline in the following sections.

\section{Pre-Optimization Steps}

\subsection{Simulatable Sewing Pattern Generation} 
From a single-view image, our pipeline starts by generating an initial sewing pattern decomposition along with stitch information using SewFormer \cite{liu2023towards}. Following SewFormer's convention, the sewing pattern is represented as a set of quadratic Bézier curves on a 2D plane, forming a collection of disconnected patches. The curves for each patch are connected to form a loop. Let $\mathcal{E}$ denote the set of all curves, with its control parameters comprising the set of curve vertices $\mathcal{P} = \{\bm P_i\}$ and the set of control points $\mathcal{K} = \{\bm K^e\}$ for each edge curve $e \in \mathcal{E}$. To enable garment simulation, the patches are discretized into triangle meshes. First, we apply arc-length parameterization to achieve uniform sampling along the patch boundaries. For stitched patch edges, we ensure they share the same number of sampled points. This consistency allows us to apply vertex-to-vertex stitch constraints in garment simulations, simplifying the sewing process. Using the sampled boundary points, we then perform Delaunay triangulation \cite{shewchuk2008two} independently for the interior of each patch.

\begin{wrapfigure}{r}{0.4\linewidth}
    \includegraphics[width=\linewidth]{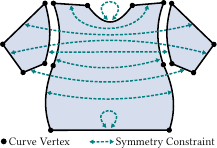}
\end{wrapfigure}
\subsubsection{Patch Symmetrization} The sewing patterns generated by SewFormer often display certain symmetries, which we aim to preserve during garment optimization. SewFormer generates a fixed number of patches with a predefined order for patch names, though some patches may remain inactive. Symmetry information, including self-symmetry and inter-symmetry, is embedded in these patch names. Symmetric edge pairs can be automatically identified by overlapping a patch with its mirrored symmetric counterpart or, in the case of self-symmetry, with the mirrored version of itself. Given the set of symmetric edge pairs $\mathcal{E}_S = \{( i, j)\sim (k, l)\}$, we define the \textit{validated} curve vertices $\{\hat{\bm P}_i\}$ of the patches prior to triangulation by solving the following quadratic optimization problem:
\begin{equation}
\min_{\{\hat{\bm P}_i\}} \sum_{( i, j)\sim (k, l)} \| (\hat{\bm P}_i - \hat{\bm P}_j) + \bm R_S(\hat{\bm P}_k - \hat{\bm P}_l)\|^2_2 + \epsilon \sum_i \|\hat{\bm P_i} - \bm P_i\|^2_2 .
\end{equation}
Here, $\bm R_S = \begin{bmatrix}-1 &0 \\ 0 & 1\end{bmatrix}$ represents the flip matrix, assuming the symmetry axis is vertical. This optimization involves solving a fixed-coefficient, positive definite linear system, which ensures differentiability. The \textit{validated} edge control points $\{\hat {\bm K}^e\}$ are computed analytically by symmetrizing their relative coordinates. The symmetrization constraints are illustrated in the inset figure. Throughout this paper, we omit the hat notation for validated vertices and control points, as all computations are based on the symmetrized patches. However, it is important to note that the underlying garment optimization variables retain the original, non-symmetry-enforced geometry parameters.

\subsubsection{Sewing Pattern Discretization} 
To enable direct optimization of Bézier curves, we make the sampling from boundary curve parameters to mesh vertices differentiable. Both boundary sampling and interior sampling are conceptualized as fixed-coordinate sampling based on their control points. Each boundary edge curve $e \in \mathcal{E}$ is defined by the starting vertex $\bm P_0^e$, the control point $\bm K^e$, and the endpoint $\bm P_1^e$ (which is also the starting point of the next edge). The curve can be differentiably parameterized as $\bm P^e(t) = (1-t)^2 \bm P_0^e + 2(1-t)t \bm K^e + t^2 \bm P_1^e$. Uniform sampling along the curve in terms of arc length is represented as a set of parameters $\{t_1^e, \ldots, t_{n_e}^e\}$, with $\bm V^{e}_i = \bm P^e(t_i^e)$ being the sampled points. The number of sampled points $n_e$ may vary for different edges. After independent triangulation for each patch, we compute the harmonic coordinate matrix $\bm H \in \mathbb{R}^{n_I\times n_B}$ \cite{joshi2007harmonic} for all the interior points, where $n_I$ is the number of interior vertices and $n_B$ is the total number of boundary vertices.
With a slight abuse of notation, we reparameterize the $j$-th interior vertex as
$\bm{V}_j^I = \sum_{i} \bm H_{ji} \bm V^B_i$, with $\bm H_{ji}$ denoting its harmonic weight relative to the $i$-th boundary point $\bm V_i^B$. Here $\bm H_{ji}$ is zero if $\bm V_j^I$ and $\bm V^B_i$ do not belong to the same patch.
During backpropagation, we fix the boundary sampling coordinates $\bigcup_{e\in \mathcal{E}, i\le n_e} \{t_i^e\}$ and the interior harmonic coordinate matrix $\bm H$, so that the triangulation is analytically determined by the original parameters of the Bézier curves. These coordinates are updated only after remeshing is performed, which will be discussed in the garment optimization section.

\subsection{Multi-view Image Generation} 
Given a single-view image of a full-body clothed human, we generate a set of multi-view RGB images and normal maps under orbital camera views using a pre-trained multi-view diffusion model, MagicMan \cite{he2024magicman}. These multi-view images of the clothed human are treated as ground truth data for human pose and garment shape in the subsequent reconstruction steps.

\subsection{Human Body Reconstruction}
The generated garment is statically draped on a fixed human mesh. To reduce the gap between the reconstructed garment and the image, an accurate human body is required to correctly support the garment. We use SMPL-X \cite{pavlakos2019expressive} as our parameterized human model. First, we apply OSX \cite{lin2023one} to the input single-view image to obtain an initial pose estimation $\bm \theta$ and shape estimation $\bm \beta$. This initial estimation typically does not perfectly align with other views, and the scaling and rotation are inconsistent across the multi-view images. Subsequently, we fine-tune the pose based on multi-view images using a coarse-to-fine strategy. 

In the coarse stage, we estimate joint landmarks on the images using DWPose \cite{yang2023effective}. Here, we optimize only the global scaling $S$ and rotation $\bm R$ of the SMPL-X model based on the following landmark loss:
\begin{equation}
\mathcal{L}_{\text{Land}}^{\text{P}} = \frac{1}{|\Omega|}\sum_{i}\|\bm w_i \cdot \left (\operatorname{Proj}(\bm J(S, \bm R, \bm \theta, \bm \beta); \Omega_i) - \bm {\bar J}_i\right)\|^2_2,
\end{equation}
where $\Omega = \{\Omega_i\}$ represents the set of camera parameters, $\bm J$ is the 3D joint location map provided by the SMPL-X model, $\operatorname{Proj}$ is the projection operator from world space to screen space, $\bm {\bar J}_i$ is the 2D joint location estimated by DWPose, and $\bm w_i$ is the per-landmark confidence score of the estimation. We use $\|\cdot\|^2_2$ to denote the mean square error (MSE). This optimization essentially estimates the model-to-world matrix of the SMPL-X model. To further refine pose and shape parameters, in the fine stage, we additionally incorporate the following RGB loss and mask loss:
\begin{align}
\mathcal{L}_{\text{RGB}}^{\text{P}} = \frac{1}{|\Omega|}\sum_{i}\| \bm (\bm M^{o}_i)^c \cdot (\bm I(S, \bm R, \bm \theta, \bm \beta, \bm C_H; \Omega_i) - \bar{\bm I}_i)\|_1, \\
\mathcal{L}_{\text{Mask}}^{\text{P}} = \frac{1}{|\Omega|}\sum_{i}\| (\bm M^{o}_i)^c \cdot (\bm M(S, \bm R, \bm \theta, \bm \beta; \Omega_i) - \bar{\bm M}_i)\|_1.
\end{align}
Here, $\bm C_H$ represents the optimizable human body vertex color, while $\bm I(\cdot)$ and $\bm M(\cdot)$ denote the posed human body RGB rendering process and contour rendering process under camera view $\Omega_i$, implemented using Nvdiffrast \cite{Laine2020diffrast}. ${\bar{\bm I}_i}$ and ${\bar{\bm M}_i}$ are the generated multi-view RGB images and masks, respectively. $\bm M^{o}_i$ represents the occluded region of the human body, which includes the garment region $\bm M^{\beta}_i$ and other non-garment occlusions $\bm M^{\alpha}_i$ (such as footwear, accessories, and hair). These regions are generated using SegFormer \cite{xie2021segformer}. The notation $(\cdot)^c$ denotes the complement of the specified region. We use $\|\cdot\|_1$ to denote the mean absolute error (MAE). By excluding the loss computation in the occluded region, we can accommodate loosely fitted garments. In summary, we optimize using the following loss in the fine stage:
\begin{equation}
\footnotesize
    \mathcal{L}^{\text{P}}(S, \bm R, \bm \theta, \bm \beta, \bm C_H) = \mathcal{L}^{\text{P}}_{\text{RGB}} + \mathcal{L}^{\text{P}}_{\text{Mask}} + \lambda_1 \mathcal{L}^{\text{P}}_{\text{Land}} + \lambda_2 \|\bm \theta - \bm \theta_0\|_1 + \lambda_3 \|\bm \beta - \bm \beta_0\|_1 .
\end{equation}
Here, we also regularize the pose and shape parameters where $\bm \theta_0$ and $\bm \beta_0$ are their initial estimates provided by OSX.

\subsection{Garment Initialization} \label{sec:init_fitting}
The generated sewing patterns are positioned near the human body and sewn together to be dressed. SewFormer provides an initial placement around the T-posed SMPL-X model. To ensure proper layering, we adopt a bottom-to-top strategy for fitting the entire set of garments onto the human body, allowing the top garments to overlay the bottom ones. Connected components are identified by treating stitched vertices as connected. These components are sorted vertically and sequentially fitted from bottom to top through simulations using CIPC. After completing the T-pose fitting, the human body is interpolated from the T-pose to the reconstructed pose, and the entire cloth-human interaction is simulated by treating the human in motion as a moving boundary condition. To secure the bottom garments and prevent them from slipping during pose interpolation, we shrink the rest shape of the triangles near the waist to generate sufficient friction.

\section{Garment Optimization}

\subsection{Optimization Overview}
In garment optimization phase, we iteratively fine tune parameters of sewing pattern so that the statically draped garments on a posed human body match generated multi-view images in all views.  We optimize the curve vertex set $\mathcal{P}$ and the control point set $\mathcal{K}$ of Bézier curves using differentiable CIPC simulation based on the generated multi-view images. To further leverage RGB information for assisting the optimization, we also optimize the vertex colors $\bm {C}_G$ of the discretized garment mesh for RGB renderings. Additionally, we optimize the global stretching stiffness $\kappa_{s}$ and the global bending stiffness $\kappa_{b}$ to automatically discover a set of physical parameters that align with the 2D observations.

For each optimization iteration, we use CIPC simulation to statically drape the garment onto the fixed-posed human body mesh. Leveraging the robustness of CIPC, we simulate one step of 1 second to directly reach near-static equilibrium. Since the static equilibrium does not locally depend on the initial state, meaning that the Jacobian matrix of the simulated state with respect to the initial state is zero, we update the initial state of iteration $n$, $\bm x_0^{n}$, to the previously simulated state:
\begin{equation}
\bm x_0^{n} = \operatorname{Sim}(\bm x_0^{n-1}; \bm \varsigma(\kappa_{s}, \kappa_{b}, \mathcal{P}, \mathcal{K})).
\end{equation}
Here, $\operatorname{Sim}$ represents the simulation process described in \autoref{sec:sim}. The initial state, $\bm x_0^{0}$, is obtained from the initial garment fitting described in \autoref{sec:init_fitting}. $\bm \varsigma(\mathcal{P}, \mathcal{K})$ denotes the simulation rest shape data, including nodal mass, per-stencil elastic stiffness, undistorted material space, and similar properties. To make the simulation as path-independent as possible, we avoid adding friction during the process. To prevent the bottom garments from slipping down, the boundary loop of the bottom component near the waist area is fixed.

In summary, we solve the following optimization problem:
\begin{equation}
\min \mathcal{L}(\mathcal{P}, \mathcal{K}, \kappa_{s}, \kappa_{b}; \bm x, \bm x_{0}),
\end{equation}
where $\bm x$ represents the simulated state starting from initial state $\bm x_{0}$, which is iteratively updated to the previously simulated state. We elaborate on the training losses in $\mathcal{L}$ that we use in the following sections. We observe that edge curvatures $\mathcal{K}$ are more sensitive than vertex positions $\mathcal{P}$. Therefore, we employ a two-stage training approach, where in the first stage, the update of 
$\mathcal{K}$ is frozen.

\subsection{Rendering Losses}
\subsubsection{Garment Mask Loss}
The dominant rendering loss we employ is the garment mask loss. Given the multi-view ground-truth images, we use SegFormer \cite{xie2021segformer} to segment top, bottom, and dress garment masks, assigning each component with a distinct color. The mask loss is defined as follows:
\begin{equation}
    \mathcal{L}_{\text{Mask}} = \frac{1}{|\Omega|}\sum_i \|(\bm M^{\alpha}_i)^c \cdot  (\bm M(\bm x; \bm C_C, \Omega_i) - \bar{\bm M}_i)\|_1.
\end{equation}
Here, $\bm x$ represents the simulated state of garments draped over the human body. $\bm C_C$ denotes the component color, which is discussed in the following section.
The rendered colored mask $\bm M(\bm x; \bm C_C, \Omega_i)$ is obtained by assigning $\bm C_C$ to the corresponding garment vertices and setting the human body to black, ensuring that only the non-occluded parts of the garments are rendered. ${\bar{\bm M}_i}$ is the set of colored garment masks generated from multi-view RGB images. We also exclude the loss computation in the occluded regions ${\bm M^{\alpha}_i}$ caused by hair and accessories to avoid incorrect mask guidance.

\paragraph{Initialization of Component Colors}
The component color $\bm C_C$ is automatically assigned prior to garment optimization. SewFormer typically predicts garments with one or two connected components. We vertically sort the sewn garment components and the 2D mask regions from the first camera view. The component colors are then assigned accordingly. If only one component is predicted but multiple garment masks are present, we adjust the multi-view garment masks to use a single color.

\subsubsection{RGB and Normal Rendering Loss}
We also utilize RGB and normal rendering losses to improve garment optimization. These losses are introduced to stabilize the training process, as the gradient of the mask rendering loss within the interior regions of the garment is zero. They are formulated similarly to the mask rendering loss:
\begin{align}
\mathcal{L}_{\text{RGB}} = \frac{1}{|\Omega|}\sum_i \|\bm M^{\beta}_i \cdot  (\bm I(\bm x; \bm C_G, \Omega_i) - \bar{\bm I}_i)\|_1, \\
\mathcal{L}_{\text{Normal}} = \frac{1}{|\Omega|}\sum_i \|\bm M^{\beta}_i \cdot  (\bm N(\bm x; \Omega_i) - \bar{\bm N}_i)\|_1.
\end{align}
Here, $\bm I(\bm x; \bm C_G, \Omega_i)$ represents the garment RGB rendering of the vertex color $\bm C_G$ under the camera view $\Omega_i$, and $\bm N(\bm x; \Omega_i)$ denotes the corresponding normal map rendering. The sets $\{\bar{\bm I}_i\}$ and $\{\bar{\bm N}_i\}$ are the multi-view RGB and normal images generated by the multi-view diffusion process. The loss computation is restricted to the garment regions $\bm M^{\beta}_i$.

\subsection{Geometric Regularizers}

The sewing pattern optimization under rendering losses alone is ill-posed because, for the same sewn 3D garment mesh, there are infinitely many ways to decompose the mesh into flattened patches. Therefore, we incorporate several geometric losses to regularize the sewing pattern optimization.

\subsubsection{Area Ratio Loss}
We use the following area ratio loss to preserve the relative area of each patch with respect to the connected component it belongs to:
\begin{equation}
\mathcal{L}_{\text{AR}} = \frac{1}{N_P} \sum_p \left(\frac{\bar{A}_p(\bm X)}{\bar{A}_p(\bm X^0)} - 1\right)^2,
\end{equation}
where $N_P$ is the number of garment patches, $\bar{A}_p$ is the operator that computes the ratio between the area of the $p$-th patch and the area of the component. $\bm X$ represents the current 2D discretization of the garment patches, and $\bm X^0$ denotes the initial sampling.

\subsubsection{Corner Regularizers}
\begin{wrapfigure}{r}{0.4\linewidth}
\vspace{-0.5em}
    \includegraphics[width=\linewidth]{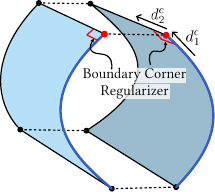}
\end{wrapfigure}
\paragraph{Boundary Corner Regularizer}
For the boundary loops of garment components, we identify all corner vertices of the original Bézier curves. At these corners, where two patches are typically sewn together, we apply the following boundary corner regularizer to penalize deviations of corner angles from right angles, as illustrated in the inset figure:
\begin{equation}
\mathcal{L}_{\text{BC}} = \frac{1}{N_{BC}}\sum_{c} (1-\bm d^c_{1} \times \bm d^c_{2}).
\end{equation}
Here, $N_{BC}$ represents the total number of boundary corners, $\bm d^c_{1}$ and $\bm d^c_{2}$ denote two consecutive unit tangent vectors at corner $c$.

\paragraph{Small-Angle Corner Regularizers}
Small angles at patch corners can introduce instabilities into optimization and simulation; thus, we use the following regularizer to penalize such angles:
\begin{equation}
\mathcal{L}_{\text{SAC}} = -\frac{1}{N_{C}} \sum_{c} s_c(\bm X) \widehat{(\bm V_1^c - \bm V_0^c)} \times \widehat{(\bm V_2^c - \bm V_0^c)}.
\end{equation}
\vspace{-0.7em}
\begin{wrapfigure}{r}{0.3\linewidth}
    \includegraphics[width=\linewidth]{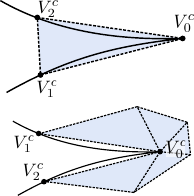}
\end{wrapfigure}
Here, $N_{C}$ is the number of patch corners, $(\bm V_1^c, \bm V_0^c, \bm V_2^c)$ is the tuple of three consecutive discrete boundary sampling points at the corner $c$, $\widehat{(\cdot)}$ is the vector normalization operator. $s_c(\bm X)$ is a non-differentiable sign function: $s_c(\bm X) = 0$ if the discretized corner angle is larger than some threshold,  otherwise, $s_c(\bm X)$ equals the sign of the cross product on the initial sewing pattern. This regularization applies to the two cases illustrated in the inset figure. It tries to maintain the same sign of the angle and avoid the angle becoming too small. However, Bézier curves may still intersect at corners even though the discretized corner triangles are normal. We use the following discretization consistency regularizer to align the curve's end tangents and discrete edge directions:
\begin{equation}
\mathcal{L}_{\text{DC}} = \frac{1}{N_{C}}  \sum_{c} (2 - \bm \tau_1^c \cdot \widehat{(\bm V_1^c - \bm V_0^c)} - \bm \tau_2^c \cdot \widehat{(\bm V_2^c - \bm V_0^c)}),
\end{equation}
where $\bm \tau_1^c, \bm \tau_2^c$ are two consecutive end tangents of Bézier curves at corner $c$.

\subsubsection{Comfort Loss}
In addition to the appearance of the fitting matching the observation, we also aim to ensure that the fitting is comfortable. We use the stretching elasticity energy to evaluate the tightness of the fitting. To prevent overly tight fitting, we introduce the following comfort regularizer:
\begin{equation}
\mathcal{L}_{\text{Comfort}} = \int \|\bm F(\bm x, \bm X) - \bm R(\bm F)\|^2 d\bm X,
\end{equation}
where $\bm R(\bm F)$ represents the closest rotation matrix to $\bm F$. This is the same as the as-rigid-as-possible (ARAP) stretching energy used in the forward simulation, except that here we assume the global stiffness is 1.

\subsubsection{Laplacian Loss}
To ensure the smoothness of the fitting, we include a Laplacian regularizer:
\begin{equation}
\mathcal{L}_{\text{Lap}} = \|\Delta \bm x\|_2,
\end{equation}
where $\Delta$ represents the node-area-weighted Laplacian operator on triangle meshes, and $\bm x$ denotes the simulated garment vertex positions.

\subsubsection{Seam Losses}
The stitched curved edge pairs should have the same shape to prevent undesired wrinkles near the seams. To achieve this, we use a seam length regularization similar to \cite{li2024diffavatar} to regularize the paired stitched edges:
\begin{equation}
\mathcal{L}_{\text{SL}} = \frac{1}{N_{S}}\sum_{e_i \sim e_j} \left |\int \|\dot{\bm P}^{e_i}(t)\| dt - \int \|\dot{\bm P}^{e_j}(t)\| dt \right |,
\end{equation}
where $N_S$ is the number of stitched seams, $e_i$ and $e_j$ iterate over all stitched edge pairs, and $\dot{\bm P}^e(t)$ represents the tangent vector. The integral is computed using finite difference and the Riemann sum. Additionally, we regularize the seam curvatures on these pairs to preserve their initial curvatures:
\begin{equation}
\mathcal{L}_{\text{SC}} = \frac{1}{2N_{S}}\sum_{e_i \sim e_j} \|\bar{\mathcal{K}}^{e_i} - \bar{\mathcal{K}}^{e_i, 0}\| + \|\bar{\mathcal{K}}^{e_j} - \bar{\mathcal{K}}^{e_j, 0}\|,
\end{equation}
where $\bar{\mathcal{K}}^e$ represents the relative coordinate of the control point within the frame of the curved edge segment $e$, and $\bar{\mathcal{K}}^{e,0}$ denotes its initial value.

\subsection{Post-Iteration Processing}
Occasionally, when two Bézier curves come close to each other—such as when forming a thin strip—the curves may penetrate one another after a parameter update in some iteration. This can lead to flipped triangles, causing the simulation to fail in the next iteration. To address this, we enforce a safeguard that modifies the geometry in-place to prevent such occurrences. Specifically, we optimize the negative triangle areas using a least-squares penalty after each iteration $n$:
\begin{equation}
    \mathcal{L}_{\text{Flip}}(\mathcal{P}, \mathcal{K}) = \frac{1}{|F|}\sum_{f} (\epsilon - \min\{A_f(\bm X), \epsilon \}) +  \lambda^{\text{Flip}} \|\bm X - \bm X^{n+1}\|_{1},
\end{equation}
where $|F|$ is the number of faces $A_f$ is the signed area of triangle $f$ and $\bm X^{n+1}$ is the discretized garment vertices after the parameter update at iteration $n$. We optimize the above loss only if triangles are close to flipping.

\begin{figure}
    \centering
    \includegraphics[width=\linewidth]{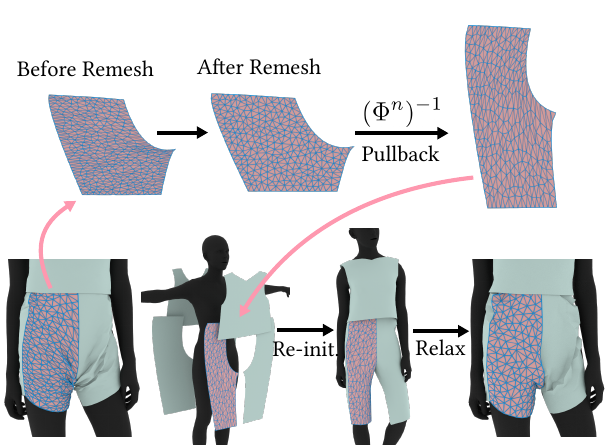}
    \caption{\textbf{Sewing Pattern Remeshing.} We perform automatic remeshing during optimization when ill-conditioned triangles are detected. To avoid penetration, we pull back the new discretization to the initial unoptimized stage and rerun the garment initialization to fit it onto the human.}
    \label{fig:remesh}
\end{figure}

\subsection{Remeshing}
During optimization, we use cage deformations defined by a fixed set of harmonic coordinates to deform a fixed number of interior vertices. The triangulation quality can degrade significantly in regions with large deformations, creating challenges for simulations. To address this, we introduce automatic remeshing during the optimization iterations when the mesh quality drops below a predefined threshold. While rerunning the discretization on updated sewing patterns is straightforward, directly remeshing the fitted garment state on the human body can lead to penetrations. This occurs because the underlying smoothly interpolated surface may intersect after re-triangulation, as the collision handling relies on the previous discretization. To resolve this, we propose a refitting procedure that sews and refits the garment patches onto the human body without causing penetrations.

Assume $\bm \chi^0$ is the initial garment sewing pattern in the continuous domain with triangulation $\mathcal{T}^0$. The sewing pattern optimization at step $n$ can be characterized by a map $\Phi^n$ from $\bm \chi^0$ to $\bm \chi^n$, where $\Phi^n$ is a piecewise linear map defined on the continuous domain. Observe that the initial fitting is sewn from the discretization of $\bm \chi^0$, where SewFormer provides reasonable transformations to position the panels around the human body. After generating a new triangulation $\mathcal{T}^n$ of $\bm \chi^n$, we pull $\mathcal{T}^n$ back to $\bm \chi^0$ as the new triangulation of $\bm \chi^0$: $\tilde{\mathcal{T}}^0 \leftarrow [{\Phi^n}]^{-1}(\mathcal{T}^n)$. We then apply the initial transformations to the updated discretization $\tilde{\mathcal{T}}^0$ to position the patches around the T-pose human body and execute the fitting procedure described in \autoref{sec:init_fitting}. During this fitting process, we set the rest shape as $\tilde{\mathcal{T}}^0$. A relaxation process follows, using $\mathcal{T}^n$ as the rest shape. The newly fitted results are non-penetrating, and we set them as the initial state $\bm x^n_0$ for the differentiable simulation process. Finally, $\mathcal{T}^0$ is replaced with $\tilde{\mathcal{T}}^0$. This remeshing process is illustrated in \autoref{fig:remesh}.

\section{Post-Optimization Steps}
\subsection{Texture Generation}

To complete our pipeline and deliver a fully textured garment directly from a single image input, tailored to the needs of the garment fabrication industry, we incorporate an additional texture generation module. Unlike formulating texture creation as a reconstruction task—an approach constrained by the ill-posed nature of the problem due to sparse inputs, severe distortion, and occlusions caused by the human body and overlapping garment layers—our module adopts generative methods to produce garment textures. This module employs two strategies for texture generation:

\paragraph{Tileable Texture Generation via FabricDiffusion}

In this strategy, we assume that in real-world garment creation, clothing panels are typically cut from a single piece of fabric and sewn together, resulting in similar and tileable textures. Based on this assumption, given the front-view ground truth input image and its corresponding colored segmentation mask, we identify the largest uniform color square area within the segmentation mask for each garment component (e.g., top or bottom) as the captured texture region. This region may exhibit distortions and varying illumination caused by occlusions and poses in the input image. To address these issues, we process the captured texture region using FabricDiffusion~\cite{zhang2024fabricdiffusion}, which generates distortion-free and tileable texture maps. To determine the appropriate tiling scale for aligning the textures with the garment's UV space (optimized in our pipeline), we assume consistent camera view parameters for the front view. This scale can be calculated by multiplying the derivative of the cropped region's size by a constant factor.

\paragraph{In-the-Wild Texture Generation via GPT-4o and FLUX}
For generalized textures that do not fall into the above case, we utilize Vision-Language Models (VLMs) in collaboration with a Diffusion model. Specifically, we process the input image using the GPT-4o~\cite{achiam2023gpt} VLM to extract descriptive keywords for the textures of various components, such as \texttt{\textcolor{blue}{"denim, dark blue, smooth fabric"}} and \texttt{\textcolor{blue}{"argyle, grey and white, knitted fabric"}} through prompt-based querying. These extracted keywords are then fed into FLUX~\cite{flux2023}, which generates the corresponding textures.

\subsection{Showcase under Human Motions}
The reconstructed simulation-ready garment and human model can be used to generate realistic dynamic human motion in clothing using the robust CIPC physics-based simulator. However, IPC-based simulators require the initial configuration of the human model to be penetration-free. As IPC-based simulators produce intersection-free results, self-penetrations of the human model during the given motion can cause solution failures when the human model interacts with garments. To address this issue, we replace the human model during motion with the nearest intersection-free human model by solving Injective Deformation Processing (IDP)~\cite{IDP} problems. In solving these IDP problems, we follow the method in \citet{garmentdreamer} where the authors use PBD simulations to resolve collisions between garments and human models. Instead, we apply extended Position-Based Dynamics (XPBD)~\cite{macklin2016xpbd} simulations to resolve self-penetrations in the human model by repelling colliding vertices and faces, while preserving natural deformations. 

\begin{figure*}[ht]
    \includegraphics[width=0.9\textwidth]{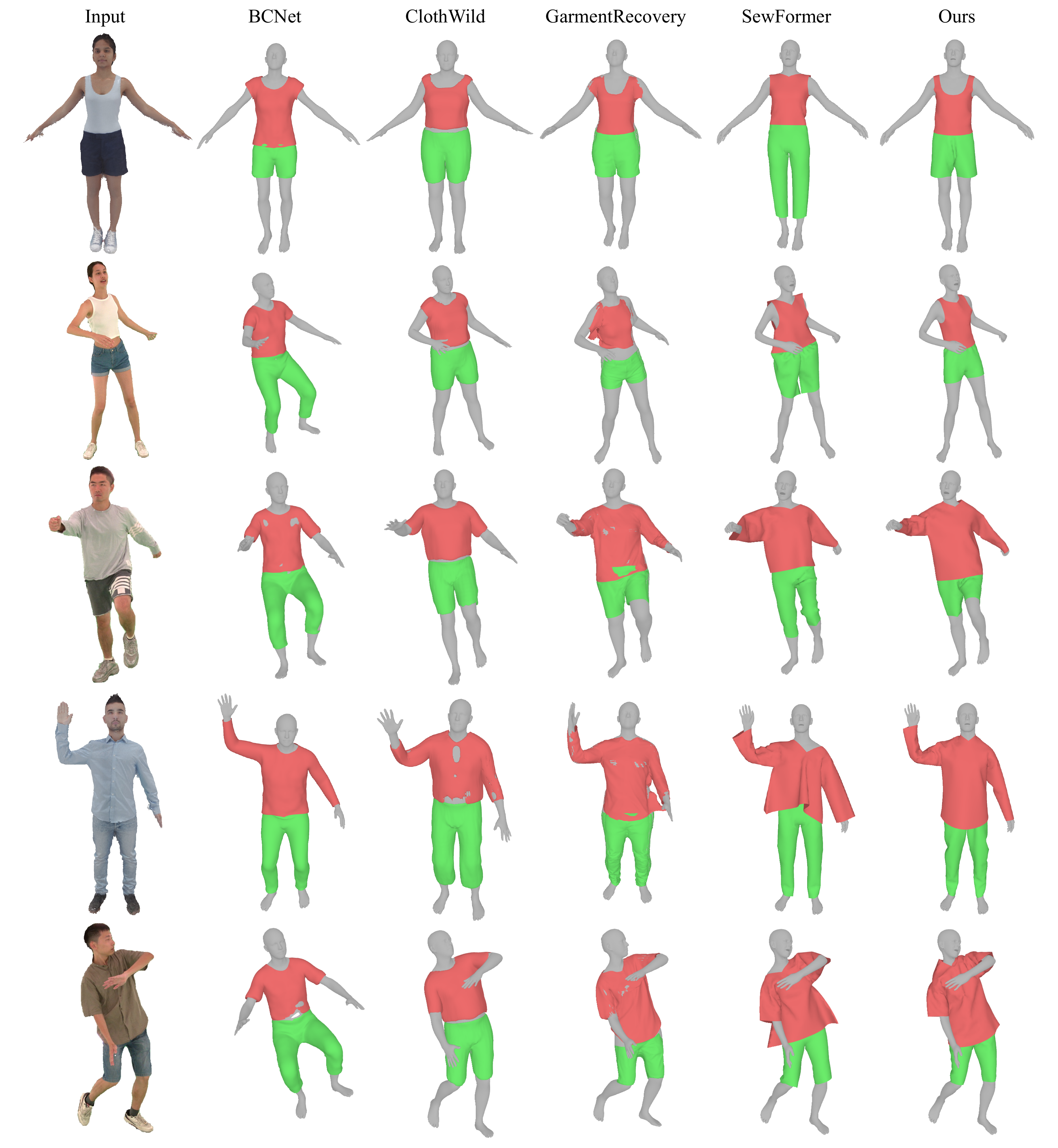}
    \centering
    \caption{\textbf{Qualitative Comparisons of Geometry Reconstruction.} Our proposed method not only generates sewing patterns that seamlessly integrate into animation and simulation workflows but also achieves superior garment reconstruction accuracy compared to baseline methods.}
    \label{fig:qualitative_comparison}
\end{figure*}

\begin{figure*}[t]
    \centering
    \includegraphics[width=\textwidth]{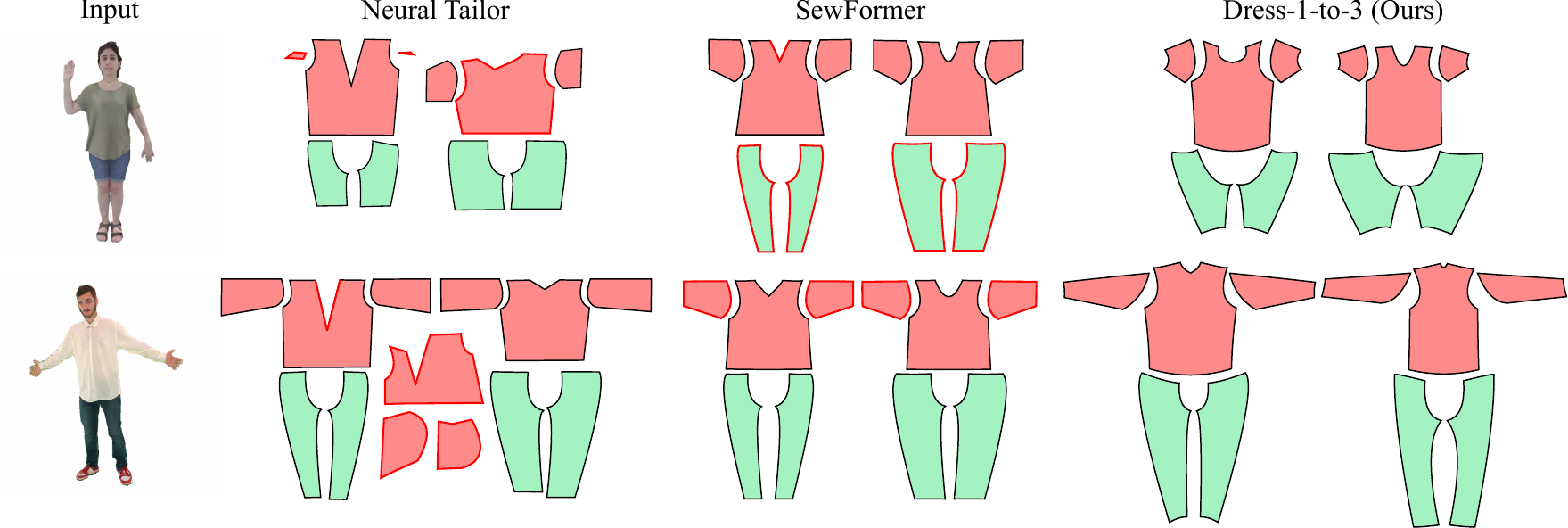}
    \caption{\textbf{Qualitative Comparison of Panel Shape Prediction.} Neural Tailor \cite{korosteleva2022neuraltailor} takes ground-truth garment meshes as input, while SewFormer \cite{liu2023towards} and our proposed method use single-view images as input. Extra unexpected panels and edges with significant errors are highlighted in red.}
    \label{fig:sewing_pattern_comparison}
\end{figure*}

\section{Implementation}
\paragraph{Differentiable Simulation Layer} We implement CIPC simulation using NVIDIA Warp \cite{macklin2022warp} to utilize the Auto-Diff feature. The simulator is wrapped in a customized \texttt{autograd.Function} to be integrated to the global computational graph. 

\paragraph{Balancing between Losses}
The training loss is weighted sum of all rendering losses and geometric regularizers. We use the $L_{\text{Mask}}$ as the dominant loss and set the relative weights for $L_{\text{RGB}}$ and $L_{\text{Normal}}$ to $\lambda_{\text{RGB}} = 0.1$ and $\lambda_{\text{Normal}} = 0.1$. For geometric regularizers, we do not have a rule of thumb to balance them. For all experiments, we use $\lambda_{\text{Lap}} = 0.001$, $\lambda_{\text{Comfort}} = 0.1$, $\lambda_{\text{AR}} = 0.01$, $\lambda_{\text{SAC}} = 0.01$, $\lambda_{\text{DC}} = 0.001$, $\lambda_{\text{BC}} = 0.001$, $\lambda_{\text{SL}} = 0.1$, $\lambda_{\text{SL}} = 0.1$.

\paragraph{Training Time}
The pre-optimization steps require about 10 minutes. The garment optimization process can be finished within 2 hours on a single RTX 3090 with 24GB device memory. 

\section{Experiments}
\subsection{Geometry Reconstruction Comparison}
\label{sec:com}
We first conduct comparison study to evaluate the reconstruction accuracy of baseline methods and our proposed approach.

\paragraph{Benchmark}
We use the CloSe \cite{antic2024close} and 4D-Dress \cite{wang20244d} datasets for the comparison study. CloSe is a large-scale 3D clothing dataset featuring detailed segmentation across diverse clothing classes. 4D-Dress offers high-quality 4D textured scans of dynamic clothed human sequences. For evaluation, we carefully select examples encompassing a variety of human body shapes, poses, and their corresponding front-view images to establish a comprehensive benchmark.

\paragraph{Baselines}
pare our method with state-of-the-art single-view garment reconstruction methods, including BCNet \cite{jiang2020bcnet}, ClothWild \cite{moon20223d}, GarmentRecovery \cite{li2024garment}, and SewFormer \cite{liu2023towards}. Among these, BCNet and ClothWild are designed for clothed human reconstruction but are limited to tight-fitting clothing and not readily adaptable for downstream tasks such as animation and simulation. GarmentRecovery extends to loose-fitting garments reconstruction by deforming predicted rest shapes to align with input images. In contrast, SewFormer predicts corresponding sewing patterns directly from images, enabling seamless integration into animation pipelines and physical simulations. Our proposed method builds upon SewFormer and incorporate differentiable simulation to refine 2D panels and physical parameters. For SewFormer and our approach, we simulate the predicted sewing patterns and use the resulting 3D garments for quantitative comparisons.

\paragraph{Results}
We evaluate the accuracy of baseline methods and our approach using two metrics: Chamfer Distance (CD) and Intersection over Union (IoU). CD quantifies the geometric similarity between reconstructed and ground-truth meshes, while IoU assesses the alignment between the garment mask of the rendered reconstruction and the input front-view images. The quantitative results for the CloSe and 4D-Dress datasets are presented in Table~\ref{tab:quantitative_comparison}, and visualized qualitative comparisons are shown in Figure~\ref{fig:qualitative_comparison}.
BCNet and ClothWild tend to produce overly smooth garment meshes, lacking fine wrinkle details. GarmentRecovery improves geometric details but often results in interpenetrated reconstructions. SewFormer predicts sewing patterns that can be directly used for simulation, yet it neglects physical parameters, leading to simulated results that deviate significantly from the ground-truth mesh. In contrast, our method not only generates sewing patterns for seamless integration into downstream pipelines but also optimizes garment physical parameters, enabling accurate geometry reconstruction that closely aligns with ground truth.

\begin{table}[t]
    \centering
    \caption{\textbf{Quantitative Comparisons of Geometry Reconstruction.} We evaluate the performance of baseline methods and our approach on the CloSe and 4D-Dress datasets. Our proposed method achieves the highest reconstruction accuracy across both datasets.}
    \label{tab:quantitative_comparison}
    \begin{tabular}{@{\extracolsep{7pt}}lcccc}
    \toprule
    \multirow{2}*{Method}   & \multicolumn{2}{c}{CloSe} & \multicolumn{2}{c}{4D-Dress} \\
    \cline{2-3} \cline{4-5}
                        &CD$\downarrow$ & IoU$\uparrow$   &CD$\downarrow$   &IoU$\uparrow$\\
    \midrule
    BCNet               &2.277   &0.781  &4.704   &0.575\\
    ClothWild           &2.166   &0.664  &3.125   &0.664\\
    GarmentRecovery     &2.058   &0.831  &2.983   &0.776\\
    \midrule
    SewFormer           &2.233   &0.748  &2.926   &0.720\\
    Dress-1-to-3 (Ours) &\textbf{1.623}   &\textbf{0.862}  &\textbf{2.441}   &\textbf{0.808}\\
    \bottomrule
    \end{tabular}
\end{table}

\subsection{Sewing Pattern Evaluation}

We compare our method with two approaches that predict garment sewing patterns: Neural Tailor \cite{korosteleva2022neuraltailor} and SewFormer \cite{liu2023towards}. Neural Tailor generates sewing patterns from garment point cloud inputs, while SewFormer and our method recover patterns directly from single-view image inputs. For comparison purposes, we sample points from garment meshes to serve as inputs for Neural Tailor. The qualitative results are shown in Figure~\ref{fig:sewing_pattern_comparison}.
Neural Tailor is trained on garments draped over an average SMPL female body in a T-pose. Consequently, its predictions are usually unsatisfactory if the human pose deviates from the T-pose, and may produce unexpected additional panels. Furthermore, its reliance on garment point clouds as input significantly restricts its practical applicability. SewFormer, on the other hand, generates symmetric and organized panels. However, its predicted panel shapes often fail to align with the input image. For instance, it may predict long pant panels for an image with short pants. This is probably due to the small scale of the dataset used for its training. Nevertheless, collecting a large-scale dataset of real-world clothed human images paired with corresponding garment meshes and sewing patterns is a challenging and resource-intensive task. In contrast, our optimization-based method requires no additional training data. By leveraging differentiable simulation, it refines an initial estimate of the sewing patterns, achieving significantly more accurate results.

\subsection{Textured Garment Reconstruction and  Simulation}
\label{sec:res}
\begin{figure*}[ht]
    \centering
    \includegraphics[width=\textwidth]{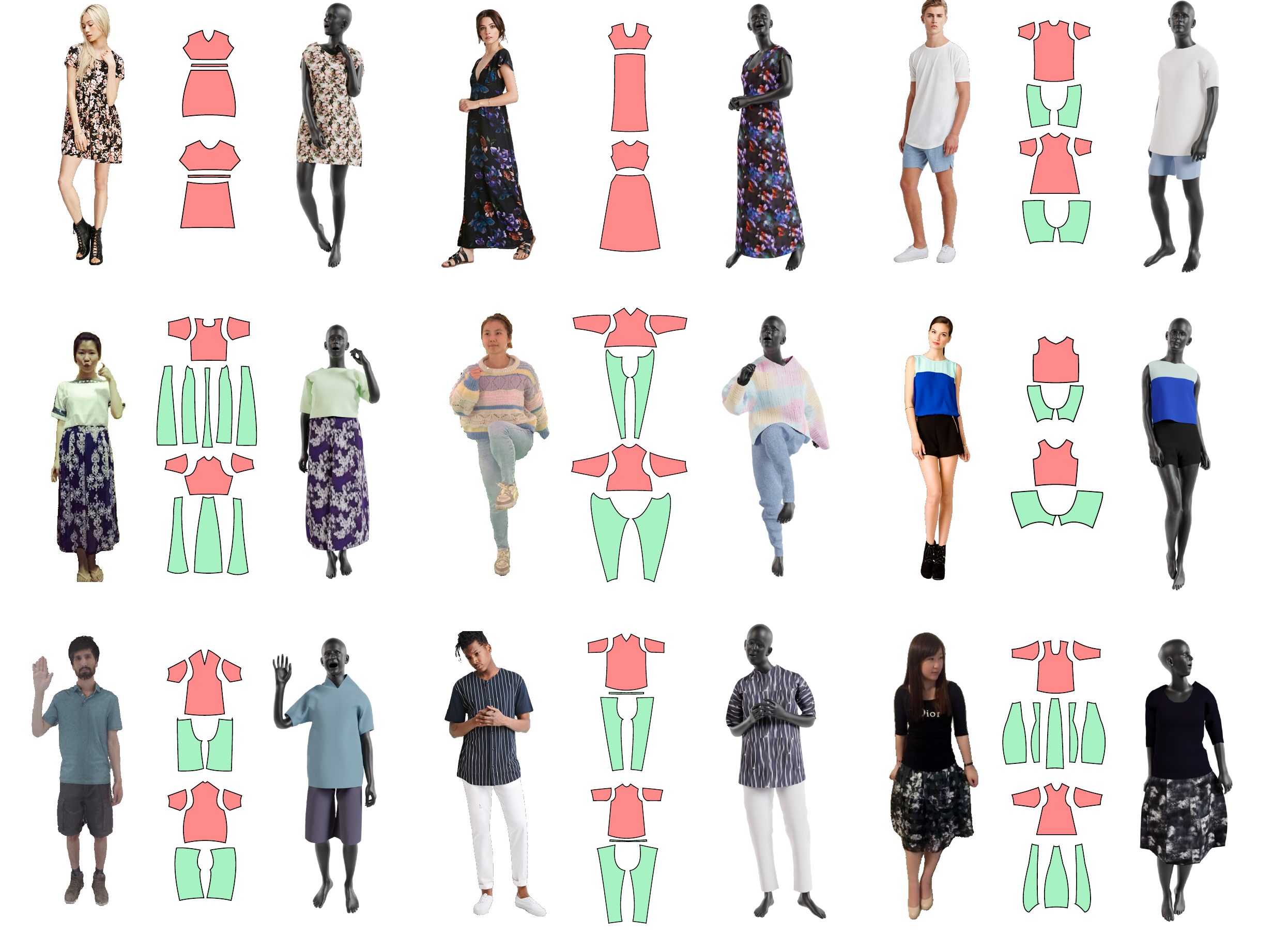}
    \caption{\textbf{Qualitative Results of Textured Clothed Human.} We showcase the generation capability of Dress-1-to-3 using in-the-wild test images from various sources, including both real-world and synthetic images. Our streamlined pipeline generates perfectly fitted 3D garments with visually plausible textures.}
    \label{fig:show_results}
\end{figure*}
\paragraph{Test Images}
To evaluate the generative capability of our method, we perform extensive tests on a variety of images from different sources including 4D-Dress \cite{wang20244d}, CloSe \cite{antic2024close} and DeepFashion2 \citep{ge2019deepfashion2}. These images exhibit diverse quality levels and human poses, highlighting the robustness of our method. To further extend the generative capability of our method with text prompts, we employ FLUX \cite{flux2023} to generate input images using custom textual descriptions of clothing on a model. For instance, prompts such as \texttt{\textcolor{blue}{"a female model wearing a blazer and pants"}} are used. To enhance the diversity of the generated results, we randomly incorporate detailed descriptions, including the shape and color of the clothing, as well as the pose and appearance of the model.

\paragraph{Textured Garment Reconstruction}
As demonstrated in \autoref{fig:show_results}, Dress-1-to-3 effectively reconstructs 3D garments that accurately fit human models in both real-world and synthetic images. Our method automatically retrieves visually plausible garment textures using image diffusion techniques. This streamlined process requires minimal human effort to reconstruct high-fidelity garments with sewing patterns and offers users the flexibility to easily adjust garment shape and texture.  

\begin{figure}
    \centering
    \includegraphics[width=\linewidth]{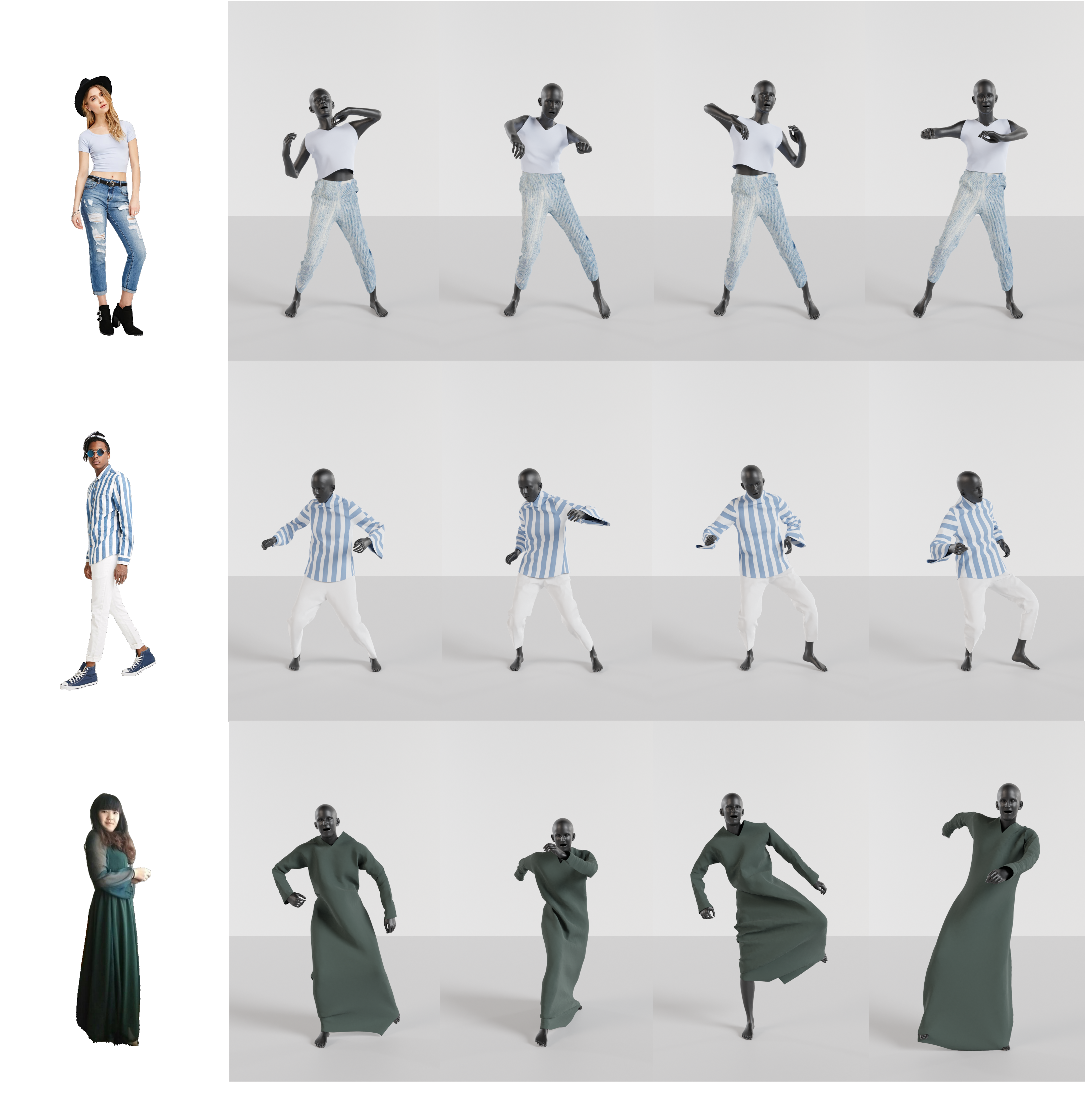}
    \caption{\textbf{Garment Simulation.} We animate garment motion using various human sequences as moving boundary conditions. Our simulation-ready garments exhibit physically plausible dynamics.}
    \label{fig:simulation}
\end{figure}

\paragraph{Garment Simulation} 
The garments synthesized by our method are simulation-ready due to the accurate sewing, fitting, and optimization of garment patterns. The optimized 3D outfits align perfectly with the human body at steady state, avoiding artifacts such as self- or interpenetration. These garments can be seamlessly integrated into physics-based simulations, such as those used in video games. In \autoref{fig:teaser} and \autoref{fig:simulation}, we visualize several simulated human motion sequences showcasing dynamic garment behavior.

\subsection{Ablation Study}
\begin{figure}
    \centering
    \includegraphics[width=\linewidth]{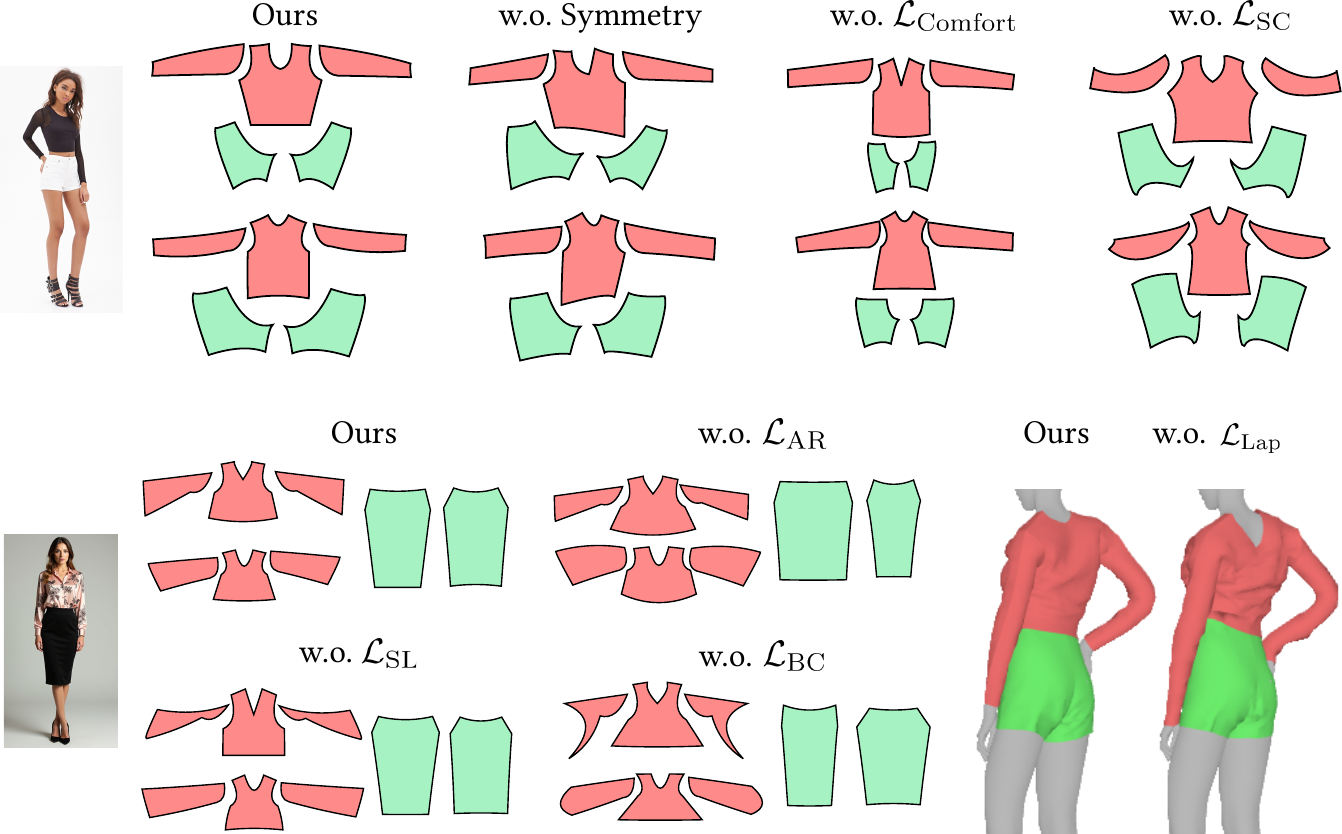}
    \caption{\textbf{Ablation Study.} We conduct ablation studies on our geometric regularizer to ensure that the sewing pattern maintains both reasonable 2D patterns and a plausible 3D fitted shape. We minimize irregularities such as asymmetry, sharp or acute angles, and inconsistent scaling of the 2D patterns while reducing noisy geometry and unrealistic wrinkles.}
    \label{fig:ablation}
\end{figure}

In \autoref{fig:ablation}, we perform an ablation study for key individual components in Dress-1-to-3, using the same garment images as in Section \ref{sec:res}. This study evaluates the contributions of each proposed component to the final garment reconstruction quality.

\paragraph{Patch Symmetrization}
We first evaluate the effectiveness of the proposed patch symmetrization, designed to facilitate better capture of symmetrical geometry. As shown in \autoref{fig:ablation}, removing symmetry enforcement results in visibly asymmetric outputs compared to the input garment image. This highlights the critical role of symmetry enforcement in preserving structural coherence and alignment, particularly for garments with strong symmetrical patterns, such as dresses or jackets. By aligning the reconstructed mesh to expected symmetrical features, this component ensures geometric fidelity.

\paragraph{Laplacian Loss}
Laplacian loss $\mathcal{L}_{\text{Lap}}$ is applied to smooth out noise and irregular wrinkles in the reconstructed garment mesh. This loss minimizes high-frequency artifacts, enabling a cleaner and more aesthetically pleasing surface. The weight of $\mathcal{L}_{\text{Lap}}$ is a tunable parameter, allowing users to control the degree of smoothness based on their preferences. As shown in \autoref{fig:ablation}, a higher weight results in smoother results but may slightly reduce detail, whereas a lower weight preserves intricate wrinkles but may retain noise.

\paragraph{Boundary Corner Regularizer}
The boundary corner regularizer, $\mathcal{L}_{\text{BC}}$, mitigates the occurrence of sharp angles in the reconstructed sewing patterns. Sharp or acute angles can lead to practical difficulties during garment fabrication, as they introduce challenges in stitching and material handling. As demonstrated in \autoref{fig:ablation}, optimization results obtained without $\mathcal{L}_{\text{BC}}$ often generate sewing patterns with acute or impractical corner geometries, whereas incorporating this regularizer results in smoother, more fabrication-friendly boundaries.

\paragraph{Comfort Loss}
Comfort loss, $\mathcal{L}_{\text{Comfort}}$, ensures the reconstructed garment mesh adheres to an appropriate scale relative to the input image. This prevents the generation of sewing patterns that are too small or tight, which could compromise wearability. Without $\mathcal{L}_{\text{Comfort}}$, as shown in \autoref{fig:ablation}, the reconstructed sewing patterns often exhibit significantly smaller dimensions than expected, leading to impractical or unrealistic results. Incorporating this loss ensures that the final garment size aligns with user expectations and real-world usability requirements

\paragraph{Area Ratio Loss}
To maintain realistic proportions between garment parts, the area ratio loss, $\mathcal{L}_{\text{AR}}$, is applied to ensure that the relative area of each patch remains consistent with the connected components, reflecting real-world fabrication principles. For instance, in a skirt, the front and back panels should have comparable areas to align with practical garment construction. As illustrated in \autoref{fig:ablation}, omitting $\mathcal{L}_{\text{AR}}$ often results in disproportionate patch sizes, such as an overly large front skirt panel compared to the back, violating fabrication norms.

\paragraph{Seam Losses}
Two seam losses: the length seam loss, $\mathcal{L}_{\text{SL}}$, and the curvature seam loss, $\mathcal{L}_{\text{SC}}$ are adopted to ensure that stitched curved edge pairs should have the same shape to prevent undesired wrinkles near the seam, and that enforces preservation of seam curvatures, respectively. As shown in \autoref{fig:ablation}, the absence of $\mathcal{L}_{\text{SL}}$ leads to uneven sleeve seams, introducing visual artifacts and potential fabrication issues. Similarly, without $\mathcal{L}_{\text{SC}}$, the seams can become overly curved, deviating significantly from the intended design. Together, these losses contribute to producing smooth and realistic seams.

\begin{figure}
    \centering
    \includegraphics[width=\linewidth]{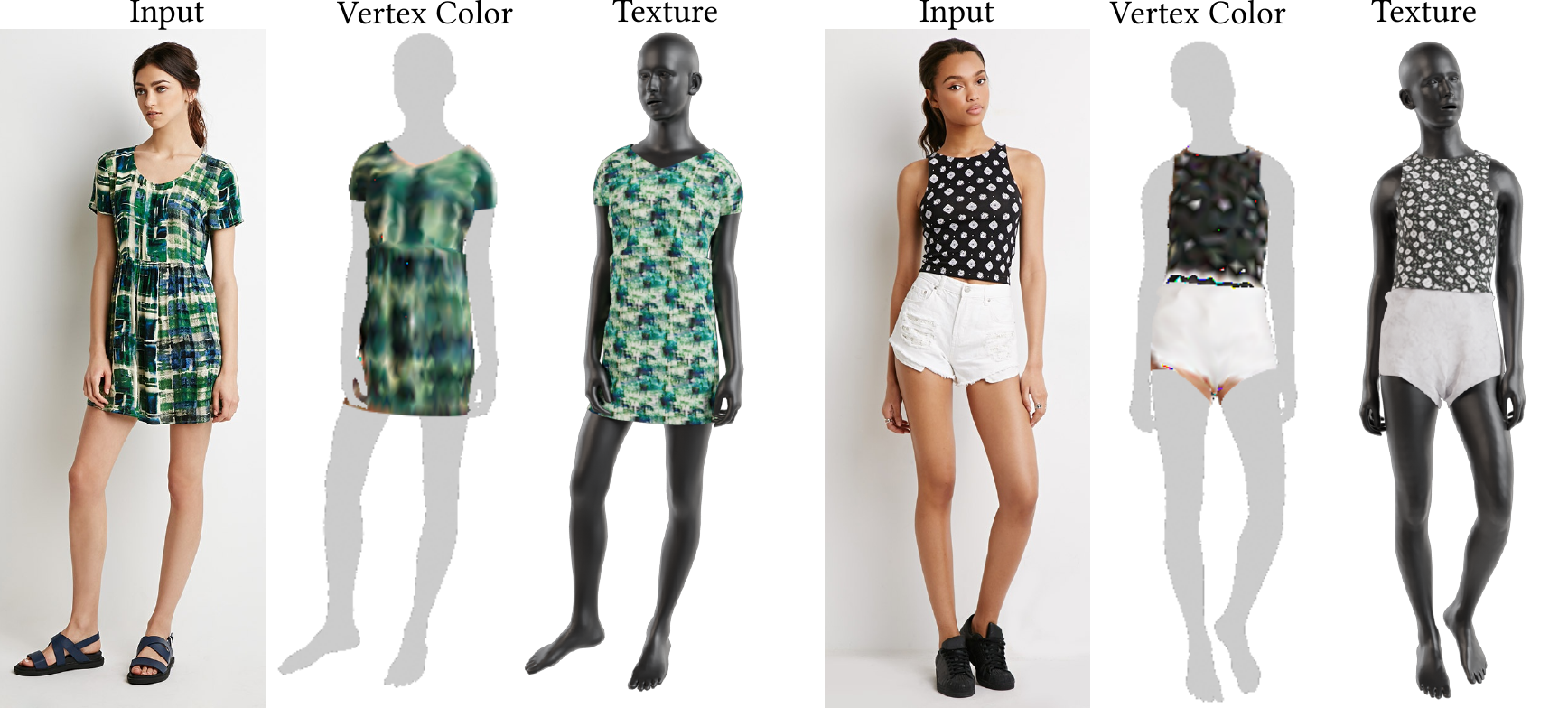}
    \caption{Comparisons between vertex color renderings and texture renderings.}
    \label{fig:vertex-color}
\end{figure}

\paragraph{Vertex Color Reconstruction}
Vertex colors are optimized to assist garment optimization. However, due to the limited mesh resolution, the visual appearance synthesized with vertex colors tends to be overly smooth. Additionally, colors from adjacent parts can bleed into part boundaries. Comparisons between vertex color renderings and texture renderings are shown in \autoref{fig:vertex-color}. This necessitates an additional module to generate textures that are not constrained by mesh resolution.

\begin{figure}
    \centering
    \includegraphics[width=\linewidth]{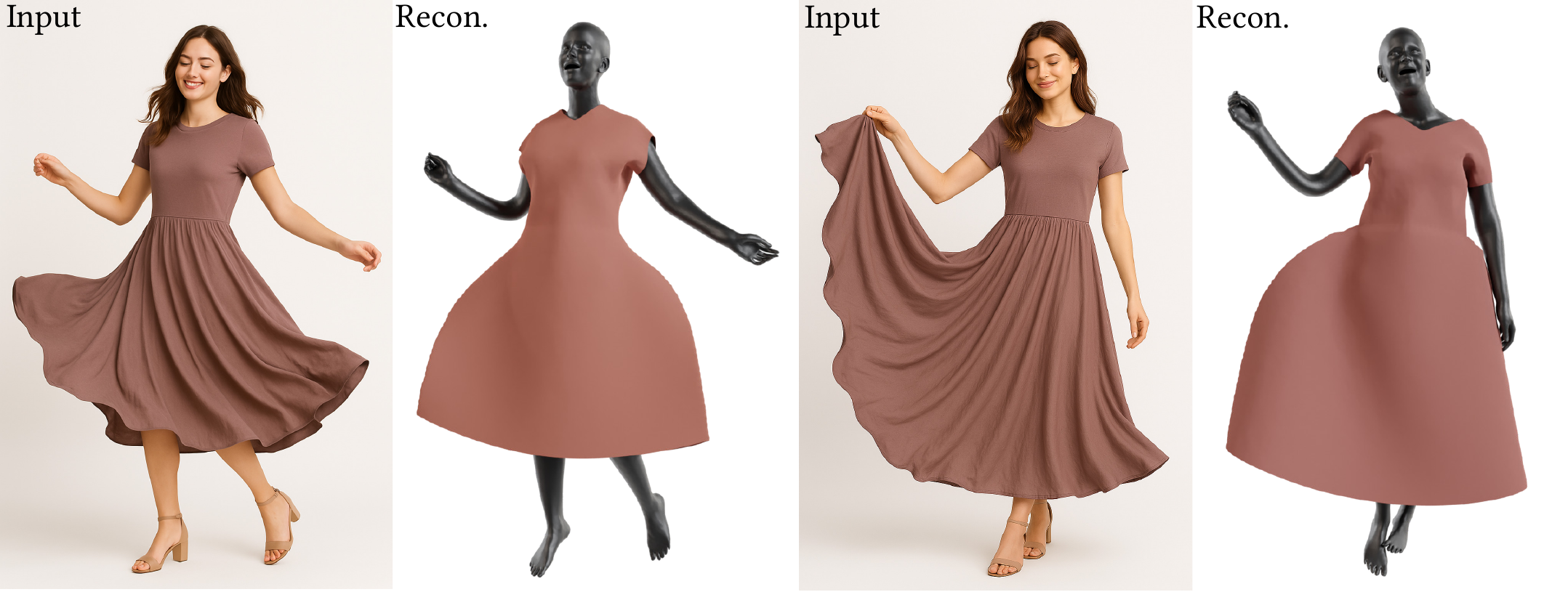}
    \caption{Our method tries to find a static garment fit that approximates a non-static garment snapshot or a fit under grasping forces. The input images are generated by GPT-4o.}
    \label{fig:loose_garment}
\end{figure}
\paragraph{Loose Garments}
Our method optimizes garments in static fitting states under gravity and body supporting forces. For non-static states, such as a snapshot of flowing, or static states influenced by other external forces such as grasping, our garment optimization process attempts to approximate the non-static states by finding nearby static configurations (as shown in \autoref{fig:loose_garment}). However, these static approximations may not reflect the garment's true geometry. We acknowledge this as a limitation and leave the extension to dynamic states or broader boundary conditions as future work.

\section{Conclusion}
In this paper, we present a garment reconstruction pipeline, Dress-1-to-3, which takes a single-view image as input and reconstructs a posed human wearing textured garments, with both the human pose and garment shapes closely aligned with the input image. During optimization, we refine the sewing pattern shapes and physical material parameters by leveraging a differentiable CIPC simulator with accurate contact. The resulting garment assets are simulation-ready and can be seamlessly integrated into a physics-based simulator.

We benchmark our pipeline against baseline methods through two key experiments: a quantitative comparison of geometry reconstruction using existing garment datasets and a qualitative evaluation of sewing patterns. In both cases, Dress-1-to-3 significantly outperforms the baseline approaches.

To further assess the Dress-1-to-3’s robustness and performance, we test our textured garment reconstruction using in-the-wild real-world and synthetic images, validated together with animations of dressed humans. The high-quality results demonstrate the robustness and effectiveness of our approach.

Additionally, ablation studies underscore the importance of the patch symmetrization technique and the contributions of each regularization loss term, highlighting their critical role in optimizing the pipeline's performance.

\paragraph{Limitations and Future Work}
While our method provides consistent high-fidelity reconstruction and has been extensively tested with in-the-wild images, its generation ability is somewhat limited by the initial estimation of the sewing pattern. For instance, our method cannot predict new connected pattern components if they are not included in the initial estimation. Additionally, challenges arise with layered clothing, as SewFormer can only predict single-layer patterns, causing multi-layered garments to be fused into a single cloth component. It is worth noting that with a more versatile sewing pattern predictor capable of handling such cases, our method would also be able to process more complex garments. We leave this enhancement as future work.

Furthermore, some optimized sewing patterns may not fully adhere to conventional fashion design principles, as our supervision relies solely on ground-truth renderings of fitted garments. Incorporating regularizers based on design conventions in future work could help produce patterns that are more suitable for  manufacturing.

Another limitation lies in the overly smoothed garment surfaces produced by our method.  To enhance training robustness, we incorporate regularizers such as seam loss and Laplacian loss; however, these also suppress the formation of natural wrinkles. Another contributing factor to the lack of high-frequency detail is the inconsistency in multi-view normal maps generated by MagicMan, which further smooths geometry in detailed regions. Addressing the reconstruction of high-frequency geometric features remains an avenue for future work. 

We also observe a gap between input images and generated textures. Since this is not our primary technical contribution, we use off-the-shelf tools for texture generation. Improving PBR texture generation is left for future work, as it warrants a standalone research effort.

Lastly, although our simulation layer supports differentiable dynamic simulation, we currently use it only for static fittings under gravity and body support forces. Extending the garment optimization to handle dynamic scenarios, such as reconstruction from monocular videos, or interaction-driven deformations like grasping, would be both interesting and practically valuable.

\paragraph{Ethical Concerns} We acknowledge that body shape biases exist in our input image datasets.

\bibliographystyle{ACM-Reference-Format}
\bibliography{reference}

\end{document}